\pdfoutput=1
\documentclass{article}

\usepackage[nonatbib,preprint]{neurips_2019}

\usepackage[utf8]{inputenc} % allow utf-8 input
\usepackage[T1]{fontenc}    % use 8-bit T1 fonts
\usepackage[bookmarks=false]{hyperref}       % hyperlinks
\usepackage{url}            % simple URL typesetting
\usepackage{booktabs}       % professional-quality tables
\usepackage{amsfonts}       % blackboard math symbols
\usepackage{nicefrac}       % compact symbols for 1/2, etc.
\usepackage{microtype}      % microtypography
\usepackage{xcolor}
\usepackage{bbm}
\usepackage{amsmath}
\usepackage{xspace}
\usepackage{graphicx}
\usepackage{multirow}
\usepackage{fdsymbol}

\newcommand{\ShortName}{LLP\xspace}

\DeclareMathOperator*{\argmax}{arg\,max}
\title{Local Label Propagation for Large-Scale Semi-Supervised Learning}

\author{%
  Chengxu Zhuang, Xuehao Ding, Divyanshu Murli, Daniel Yamins \\
  Stanford University \\
}

\begin{document}

\maketitle

\begin{abstract}
A significant issue in training deep neural networks to solve supervised learning tasks is the need for large numbers of labelled datapoints.  
The goal of semi-supervised learning is to leverage ubiquitous unlabelled data, together with small quantities of labelled data, to achieve high task performance.  
Though substantial recent progress has been made in developing semi-supervised algorithms that are effective for comparatively small datasets, many of these techniques do not scale readily to the large (unlaballed) datasets characteristic of real-world applications.
In this paper we introduce a novel approach to scalable semi-supervised learning, called Local Label Propagation (LLP).
Extending ideas from recent work on unsupervised embedding learning, LLP first embeds datapoints, labelled and otherwise, in a common latent space using a deep neural network. 
It then propagates pseudolabels from known to unknown datapoints in a manner that depends on the local geometry of the embedding, taking into account both inter-point distance and local data density as a weighting on propagation likelihood.
The parameters of the deep embedding are then trained to simultaneously maximize pseudolabel categorization performance as well as a metric of the clustering of datapoints within each psuedo-label group, iteratively alternating stages of network training and label propagation.
We illustrate the utility of the LLP method on the ImageNet dataset, achieving results that outperform previous state-of-the-art scalable semi-supervised learning algorithms by large margins, consistently across a wide variety of training regimes.
We also show that the feature representation learned with LLP transfers well to scene recognition in the Places 205 dataset.
\end{abstract}

\section{Introduction} \label{sec:intro}
Deep neural networks (DNNs) have achieved impressive performance on tasks across a variety of domains, including vision~\cite{krizhevsky2012imagenet, simonyan2014very, he2016deep, he2017mask}, speech recognition~\cite{hinton2012deep,hannun2014deep,deng2013new,noda2015audio}, and natural language processing~\cite{young2018recent, hirschberg2015advances, conneau2016very, kumar2016ask}. 
However, these achievements often heavily rely on large-scale labelled datasets, requiring burdensome and expensive annotation efforts.
This problem is especially acute in specialized domains such as medical image processing, where annotation may involve performing an invasive process on patients. 
To avoid the need for large numbers of labels in training DNNs, researchers have proposed unsupervised methods that operate solely with ubiquitously available unlabeled data. Such methods have attained significant recent progress in the visual domain, where  state-of-the-art unsupervised learning algorithms have begun to rival their supervised counterparts on large-scale transfer learning tests~\cite{caron2018deep, wu2018unsupervised, zhuang2019local, donahue2016adversarial, zhang2016colorful, noroozi2016unsupervised, noroozi2017representation}. 

While task-generic unsupervised methods may provide good starting points for feature learning, they must be adapted with at least some labelled data to solve any specific desired target task. 
%"Adapted" is confusing here -- you might mean you use labeled data for transfer learning or some training data have labels but some do not.
Semi-supervised learning seeks to leverage limited amounts of labelled data, in conjunction with extensive unlabelled data, to bridge the gap between the unsupervised and fully supervised cases. Recent work in semi-supervised learning has shown significant promise~\cite{liu2018deep,iscen2019label,zhai2019s4l,miyato2018virtual,tarvainen2017mean,lee2013pseudo,grandvalet2005semi,qiao2018deep}, although gaps to supervised performance levels still remain significant, especially in large-scale datasets where very few labels are available.  
An important consideration is that many recently proposed semi-supervised methods rely on techniques whose efficiency scales poorly with dataset size and thus cannot be readily applied to many real-world machine learning problems~\cite{liu2018deep,iscen2019label}.

Here, we propose a novel semi-supervised learning algorithm that is specifically adapted for use with large sparsely-labelled datasets. 
This algorithm, termed Local Label Propagation (\ShortName), learns a nonlinear embedding of the input data, and exploits the local geometric structure of the latent embedding space to help infer useful pseudo-labels for unlabelled datapoints.
\ShortName borrows the framework of non-parametric embedding learning, which has recently shown utility in unsupervised learning~\cite{wu2018unsupervised, zhuang2019local}, to first train a deep neural network that embeds labelled and unlabelled examples into a lower-dimensional latent space.
\ShortName then propagates labels from known examples to unknown datapoints, weighting the likelihood of propagation by a factor involving the local density of known examples. 
The neural network embedding is then optimized to categorize all datapoints according to their pseudo-labels (with stronger emphasis on true known labels), while simultaneously encouraging datapoints sharing the same (pseudo-)labels to aggregate in the latent embedding space. 
The resulting embedding thus gathers both labelled images within the same class and unlabelled images sharing statistical similarities with the labelled ones.
Through iteratively applying the propagation and network training steps, the \ShortName algorithm builds a good underlying representation for supporting downstream tasks, and trains an accurate classifier for the specific desired task.

We apply the \ShortName procedure in the context of object categorization in the ImageNet dataset~\cite{deng2009imagenet}, learning a high-performing network while discarding most of the known labels. 
The \ShortName procedure substantially outperforms previous state-of-the-art semi-supervised algorithms that are sufficiently scalable that they can be applied to ImageNet~\cite{zhai2019s4l,miyato2018virtual,tarvainen2017mean,lee2013pseudo,grandvalet2005semi,qiao2018deep}, with gains that are consistent across a wide variety of training regimes.
\ShortName-trained features also support improved transfer to Places205, a large-scale scene-recognition task. 
In the sections that follow, we first discuss related literature (\S\ref{sec:relat}), describe the \ShortName method (\S\ref{sec:method}), show experimental results (\S\ref{sec:results}), and
present analyses that provide insights into the learning procedure and justification of key parameter choices (\S\ref{sec:analysis}). 

\section{Related Work} \label{sec:relat}
Below we describe conceptual relationships between our work and recent related approaches, and identify relevant major alternatives for comparison.

\textbf{Deep Label Propagation.} 
Like \ShortName, Deep Label Propagation~\cite{iscen2019label} (DLP) also iterates between steps of  label propagation and neural network optimization.
In contrast to \ShortName, the DLP label propagation scheme is based on computing pairwise similarity matrices of learned visual features across all (unlabelled) examples.
Unlike in \ShortName, the DLP loss function is simply classification with respect to pseudo-labels, without any additional aggregation terms ensuring that the pseudo-labelled and true-labelled points have similar statistical structure. 
The DLP method is effective on comparatively small datasets, such as CIFAR10 and Mini-ImageNet. 
However, DLP is challenging to apply to large-scale datasets such as ImageNet, since its label propagation method is $O(N^2)$ in the number $N$ of datapoints, and is not readily parallelizable.
In contrast, \ShortName is $O(NM)$, where $M$ is the number of labelled datapoints, and is easily parallelized, making its effective complexity $O(NM/P)$, where $P$ is the number of parallel processes.
In addition, DLP uniformly propagates labels across networks' implied embedding space, while \ShortName's use of local density-driven propagation weights specifically exploits the geometric structure in the learned embedding space, improving pseudo-label inference.

\textbf{Deep Metric Transfer and Pseudolabels.} 
The Deep Metric Transfer~\cite{liu2018deep} (DMT) and Pseudolabels~\cite{lee2013pseudo} methods both use non-iterative two-stage procedures. 
In the first stage, the representation is initialized either with a self-supervised task such as non-parametric instance recognition (DMT), or via direct supervision on the known labels (Pseudolabels). 
In the second stage, pseudo-labels are obtained either by applying a label propagation algorithm (DMT) or naively from the pre-trained classifier (Pseudolabels), and these are then used to fine-tune the network.
As in DLP, the label propagation algorithm used by DMT cannot be applied to large-scale datasets, and does not specifically exploit local statistical features of the learned representation.
While more scalable, the Pseudolabels approach achieves comparatively poor results.
A key point of contrast between \ShortName and the two-stage methods is that in \ShortName, the representation learning and label propagation processes interact via the iterative training process, an important driver of \ShortName's improvements. 

\textbf{Self-Supervised Semi-Supervised Learning.} %S$^4$L
Self-Supervised Semi-Supervised Learning~\cite{zhai2019s4l} (S$^4$L) co-trains a network using self-supervised methods on unlabelled images and traditional classification loss on labelled images.
Unlike \ShortName, S$^4$L simply ``copies'' self-supervised learning tasks as parallel co-training loss branches.
In contrast, \ShortName involves a nontrivial interaction between known and unknown labels via label propagation and the combination of categorization and aggregation terms in the shared loss function, both factors that are important for improved performance.

\textbf{Consistency-based regularization.}
Several recent semi-supervised methods rely on data-consistency regularizations.
Virtual Adversarial Training (VAT)~\cite{miyato2018virtual} adds small input perturbations, requiring outputs to be robust to this perturbation.
Mean Teacher (MT)~\cite{tarvainen2017mean} requires the learned representation to be similar to its exponential moving average during training.
Deep Co-Training (DCT)~\cite{qiao2018deep} requires the outputs of two views of the same image to be similar, while ensuring outputs vary widely using adversarial pairs.
These methods all use unlabeled data in a ``point-wise'' fashion, applying the proposed consistency metric separately on each.
They thus differ significantly from \ShortName, or indeed any method that explicitly relates unlabelled to labelled points.
\ShortName benefits from training a shared embedding space that aggregates statistically similar unlabelled datapoints together with labelled (putative) counterparts.
As a result, increasing the number of unlabelled images consistently increases the performance of \ShortName, unlike for the Mean-Teacher method.

\section{Methods} \label{sec:method}
\begin{figure*}
\begin{center}
\includegraphics[width=\textwidth] {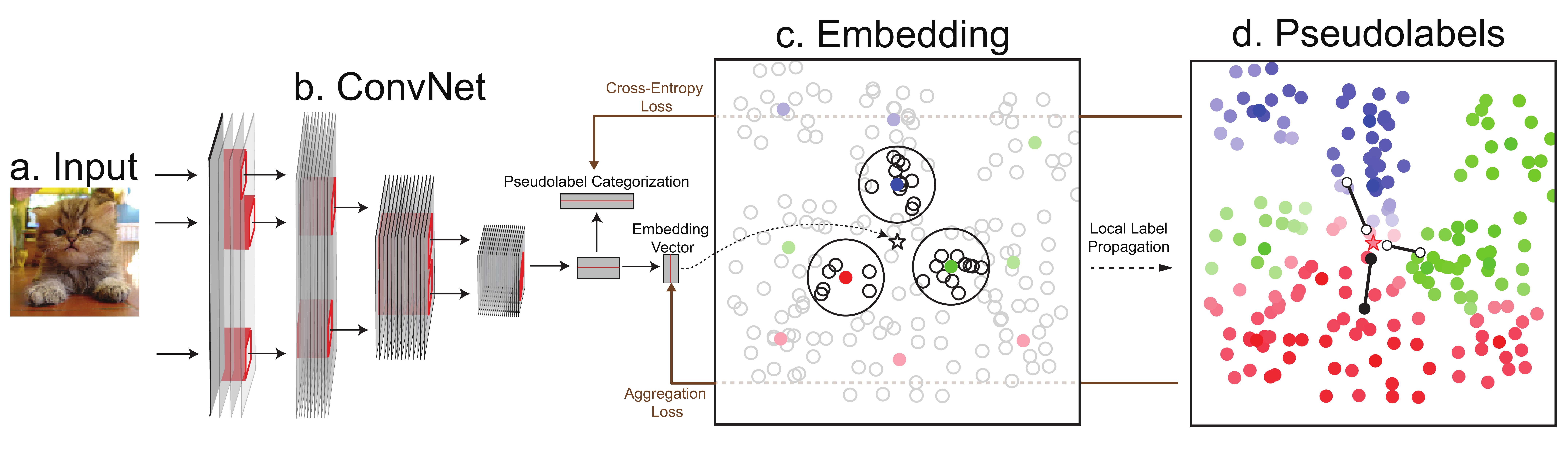}
\end{center}
\caption{
\textbf{Schematic of the Local Label Propagation (\ShortName) method.}
\textbf{a.-b.} We use deep convolutional neural networks to simultaneously generate a lower-dimensional embedding and a category prediction for each input example.
\textbf{c.} If the embedding of the input (denoted by $\star$) is unlabelled, we identify its close labelled neighbors (colored points), and infer $\star$'s pseudo-label by votes from these neighbors, with voting weights jointly determined by their distances from $\star$ and the density of their local neighborhoods (the highlighted circular areas).
\textbf{d.} The pseudolabels thereby created (colored points) come equipped with a confidence (color brightness),  measuring how accurate the pseudo-label is likely to be.
The network in \textbf{b.} is optimized (with per-example confidence weightings) so that its category predictions match the pseudo-labels, while its embedding is attracted ($\leftrightblackspoon$) toward other embeddings sharing the same pseudo-labels  and repelled ($\leftrightspoon$) by embeddings of other pseudo-labels.
}
\label{fig:schema}
\end{figure*}

We first give an overview of the \ShortName method. At a high level, \ShortName learns a model $f_\theta(\cdot)$ from labeled examples $X_L=\{x_1, \ldots, x_M\}$, their associated labels $Y_L=\{y_1, \ldots, y_M\}$, and unlabelled examples $X_U=\{x_{M+1}, \ldots, x_{N}\}$. $f_\theta(\cdot)$ is realized via a deep neural network whose parameters $\theta$ are network weights.
For each input $x$, $f_\theta(x)$ generates two outputs (Fig.~\ref{fig:schema}): an ``embedding output'', realized as a vector $v$ in a $D$-dimensional sphere, and a category prediction output $\hat{y}$. 
In learning $f_\theta(\cdot)$, the \ShortName procedure repeatedly alternates between two steps: \emph{label propagation} and \emph{representation learning}.  First, known labels $Y_L$ are propagated from $X_L$ to $X_U$, creating pseudo-labels $Y_U=\{y_{M+1}, \ldots, y_{N}\}$. Then, network parameters $\theta$ are updated to minimize a loss function balancing category prediction accuracy evaluated on the $\hat{y}$ outputs, and a metric of statistical consistency evaluated on the $v$ outputs.

In addition to pseudo-labels, the label propagation step also generates [0,1]-valued \emph{confidence scores} $c_i$ for each example $x_i$.
For labelled points, confidence scores $C_L=\{c_1, \ldots, c_M\}$ are automatically set to 1, while for pseudo-labelled examples, confidence scores $C_U=\{c_{M+1}, c_{M+2}, ..., c_{N}\}$ are computed from the local geometric structure of the embedded points, reflecting how close the embedding vectors of the pseudo-labelled points are to those of their putative labelled counterparts.
The confidence values are then used as loss weights during representation learning. 

\textbf{Representation Learning.}
Assume that datapoints $X = X_U \cup X_L$, labels and pseudolabels $Y = Y_U \cup Y_L$, and confidences $C=C_U \cup C_L$ are given.
Let $V = \{v_1, \ldots, v_N\}$ denote the set of corresponding embedded vectors, and $\hat{Y} = \{\hat{y}_1, \ldots, \hat{y}_N\}$ denote the set of corresponding category prediction outputs.
In the representation learning step, we update the network embedding parameters by simultaneously minimizing the standard cross-entropy loss $L_C(Y, \hat{Y})$ between predicted and propagated pseudolabels, while maximizing a global aggregation metric $L_A (V|Y)$ to enforce overall consistency between known labels and pseudolabels.

The definition of $L_A (V|Y)$ is based on the non-parametric softmax operation proposed by Wu \emph{et. al.}~\cite{wu2018unsupervised, wu2018improving}, which defines the probability that an arbitrary embedding vector $v$ is recognized as the $i$-th example as:
\begin{equation}
P(i|v)=\frac{exp(v_i^Tv/\tau)}{\sum_{j=1}^Nexp(v_j^Tv/\tau)},
\label{equ:softmax}
\end{equation}
where temperature $\tau\in [0,1]$ is a fixed hyperparameter.
For $S \subset X$, the probability of a given $v$ being recognized as an element of $S$ is:
\begin{equation}
    P(S|v)=\sum_{i \in S} P(i|v).
    \label{equ:sum_softmax}
\end{equation}
We then define the aggregation metric as the (negative) log likelihood that $v$ will be recognized as a member of the set of examples sharing its pseudo-label: 
\begin{equation}
  L_A (v) = -\mathrm{log}(P(A|v)), \quad \mathrm{where} \quad A = \{x_i | y_i = y\}.
  \label{equ:loss_a}
\end{equation}
Optimizing $L_A (v)$ encourages the embedding corresponding to a given datapoint to selectively become close to embeddings of other datapoints with the same pseudo-label (Fig.~\ref{fig:schema}).

The cross-entropy and aggregation loss terms are scaled on a per-example basis by the confidence score, and an $L2$ weight regularization penalty is added.
Thus, the final loss for example $x$ is:
\begin{equation}
  \mathcal{L}(x|\theta) = c \cdot [L_C (y, \hat{y}) + L_A(v)] + \lambda \| \theta \|_2^2
  \label{equ:loss_final}
\end{equation}
where $\lambda$ is a regularization hyperparameter.

\textbf{Label Propagation.} 
We now describe how \ShortName generates pseudo-labels $Y_U$ and confidence scores $C_U$.
To understand our actual procedure, it is useful to start from the weighted K-Nearest-Neighbor classification algorithm~\cite{wu2018unsupervised}, in which a ``vote'' is obtained from the top $K$ nearest labelled examples for each unlabelled example $x$, denoted $\mathcal{N}_K(x)$.
The vote of each $i \in \mathcal{N}_K(x)$ is weighted by the corresponding probabilities $P(i|v)$ that $v$ will be identified as example $i$.
Assuming $Q$ classes, the total weight for pseudo-labelled $v$ as class $j$ is thus:
\begin{equation}
  w_j(v) = \sum_{i \in I^{(j)}} P(i|v), \quad \mathrm{where} \quad I^{(j)} = \{i|x_i \in \mathcal{N}_K(v), y_i=j\}
  \label{equ:naive_weight}
\end{equation}
Therefore, the probability $p_j(v)$ that datapoint $x$ is of class $j$, the associated inferred pseudo-label $y$, and the corresponding confidence $c$, may be defined as:
\begin{equation}
  p_j(v) = w_j(v)/\sum_{k=1}^{Q} w_k(v); \quad y = \argmax_j p_j(v); \quad c = p_y(v).
  \label{equ:prob_label_conf}
\end{equation}
Although intuitive, weighted-KNN ignores all the other \emph{unlabelled} examples in the embedding space when inferring $y$ and $c$.
Fig.~\ref{fig:schema}c depicts a scenario where this can be problematic: the embedded point $\star$ is near examples of three different known classes (red, green and blue points) that are of similar distance to $\star$, but each having different local data densities.
If we directly use $P(i|v)$ as the weights for these three labelled neighbors, they will contribute similar weights when calculating $y$.
However, we should expect a \emph{higher weight} from the lower-density red neighbor. 
The lower density indicates that the prior that any given point near $v$ is identified as the red instance is lower than for the other possible labels. 
However, this information is not reflected in $P(\text{red}|v)$, which represents the joint probability that $v$ is red.
 We should instead use the \emph{posterior} probability as the vote weight, which by Bayes' theorem means that $P(\text{red}|v)$ should divided by its prior. 
To formalize this reasoning, we replace $P(i|v)$ in the definition of $w_j(v)$ with a locally weighted probability:
\begin{equation}
  P^L(i|v) = P(i|v) / \rho(v_i) \text{ where } \rho(v_i) = \sum_{j \in \mathcal{N}_T(v_i)} P(j|v_i)
  \label{equ:local_weight}
\end{equation}
where $\mathcal{N}_T(v_i)$ are $T$ nearest neighbors and denominator $\rho(v_i)$ is a measure of the local embedding density.
For consistency, we replace $\mathcal{N}_K(v)$ with $\mathcal{N}_K^L(v)$, which contains the $K$ labelled neighbors with highest locally-weighted probability, to ensure that the votes come from the most relevant labelled examples.
The final form of the \ShortName propagation weight equation is thus:
\begin{equation}
  w_j(v) = \sum_{i \in I^{(j)}} \frac{P(i|v)}{\sum_{j \in \mathcal{N}_T(v_i)} P(j|v)}, \mathrm{where}~I^{(j)} = \{i|i \in \mathcal{N}^L_K(v), y_i=j\}
  \label{equ:modified_weight}
\end{equation}

\textbf{Memory Bank.}
Both the label propagation and representation learning steps implicitly require access to all the embedded vectors $V$ at every computational step.
However, recomputing $V$ rapidly becomes intractable as dataset size increases. 
We address this issue by approximating realtime $V$ with a memory bank $\bar{V}$ that keeps a running average of the embeddings.
As this procedure is directly taken from~\cite{wu2018unsupervised, wu2018improving, zhuang2019local}, we refer readers to these works for a detailed description.

\section{Results} \label{sec:results}
%(Basic setting: architecture, learning rate schedule, optimizer, loading from pretrained model, how the classifier is trained, etc)
We first evaluate the \ShortName method on visual object categorization in the large-scale ImageNet dataset~\cite{deng2009imagenet}, under a variety of training regimes.
We also illustrate transfer learning to Places 205~\cite{zhou2014learning}, a large-scale scene-recognition dataset.

\textbf{Experimental settings.}
Following~\cite{wu2018unsupervised, zhuang2019local}, $\tau = 0.07$ and $D = 128$.
Optimization uses SGD with momentum of 0.9, batch size of 128, and weight-decay parameter $\lambda=0.0001$.
Learning rate is initialized to 0.03 and then dropped by a factor of 10 whenever validation performance saturates. 
Depending on the training regime specifics (how many labelled and unlabelled examples), training takes 200-400 epochs, comprising three learning rate drops.
Similarly to~\cite{zhuang2019local}, we initialize networks with the IR loss function in a completely unsupervised fashion for 10 epochs, and then switch to the \ShortName method.
Most hyperparameters are directly taken from~\cite{wu2018unsupervised, wu2018improving, zhuang2019local}, although as shown in~\cite{zhai2019s4l}, a hyperparameter search can potentially improve performance.
In the label propagation stage, we set $K=10$ and $T=25$ (these choices are justified in Section~\ref{sec:analysis}).
The density estimate $\rho(v_i)$ is recomputed for all labelled images at once at the end of every epoch.
For each traning regmine, we train both ResNet-18v2 and ResNet-50v2~\cite{he2016identity} architectures, with an additional fully connected layer added alongside the standard softmax categorization layer to generate the embedding output.
%\subsection{ImageNet with 10\% labels}
%\input{sections/tables/res18_res50.tex}

\textbf{ImageNet with varying training regimes.}
We train on ImageNet with $p\%$ labels and $q\%$ total images available, meaning that $M \sim p\% \times 1.2\mathrm{M}$, $N \sim q\% \times 1.2\mathrm{M}$.
Different regimes are defined by $p \in \{1, 3, 5, 10\}$ and $q \in \{30, 70, 100\}$.
Results for each regime are shown in Tables~\ref{tab:part_labels_18}-\ref{tab:part_unlabeled}.
Due to the inconsistency of reporting metrics across different papers, we alternate between comparing top1 and top5 accuracy, depending on which metric was reported in the relevant previous work.

The results show that: 
\textbf{1.} \ShortName significantly outperforms previous state-of-the-art methods by large margins within all training regimes tested, regardless of network architectures used, number $M$ of labels, and number $N$ of available unlabelled images; 
\textbf{2.} \ShortName shows especially large improvements to other methods when only small number of labels are known. For example, ResNet-18 trained using \ShortName with only 3\% labels achieves 53.24\% top1 accuracy, which is 12.43\% better than Mean Teacher; 
\textbf{3.} Unlike Mean Teacher, where the number of unlabelled images appears essentially irrelevant, \ShortName consistently benefits from additional unlabelled images (see Table~\ref{tab:part_unlabeled}), and is not yet saturated using all the images in the ImageNet set.
This suggests that with additional unlabelled data, \ShortName could potentially achieve even better performance.
\begin{table}
\begin{center}
\begin{tabular}{c|c|c|c|c}
\hline
Method & 1\% labels & 3\% labels & 5\% labels & 10\% labels \\
\hline\hline
Supervised & 17.35 & 28.61 & 36.01 & 47.89 \\
\hline
DCT~\cite{qiao2018deep} & -- & -- & -- & 53.50 \\ 
MT~\cite{tarvainen2017mean} & 16.91 & 40.81 & 48.34 & 56.70 \\
\ShortName (ours) & \textbf{27.14} & \textbf{53.24} & \textbf{57.04} & \textbf{61.51} \\
\hline
\end{tabular}
\end{center}
\caption{
ResNet-18 Top-1 accuracy (\%) on the ImageNet validation set trained on ImageNet with 10\%, 5\%, 3\%, or 1\% of labels.
Mean Teacher performance is generated by us.
}
\label{tab:part_labels_18}
\vspace{-2mm}
\end{table}
\begin{table}
\begin{center}
\begin{tabular}{c|ccccccc}
\hline
\# labels & Supervised & Pseudolabels & VAT & VAT-EM~\cite{grandvalet2005semi} & $S^4L$* & MT & \ShortName (ours) \\
\hline\hline
1\% & 48.43 & 51.56 & 44.05 & 46.96 & 53.37 & 40.54 & \textbf{61.89} \\
\hline
10\% & 80.43 & 82.41 & 82.78 & 83.39 & 83.82 & 85.42 & \textbf{88.53} \\
\hline
\end{tabular}
\end{center}
\caption{
ResNet-50 Top-5 accuracy (\%) on the ImageNet validation set trained on ImageNet using 10\% or 1\% of labels.
The numbers for models except ours and Mean Teacher's are from~\cite{zhai2019s4l}. 
Because~\cite{zhai2019s4l} only reports top-5 accuracies for these models, we also report top-5 accuracy here to be comparable to theirs.
*: for $S^4L$, we list their $S^4L$-Rotation performance, which is their best reported performance using ResNet-50.
Note that although a model with higher performance is reported by $S^4L$, that model uses a much more complex architecture than ResNet-50.
}
\label{tab:part_labels_50}
\vspace{-2mm}
\end{table}
\begin{table}
\begin{center}
\begin{tabular}{c|c|c|c}
\hline
Method & 30\% unlabeled & 70\% unlabeled & 100\% unlabeled \\
\hline\hline
MT & 56.07 & 55.59 & 55.65 \\
\ShortName (ours) & \textbf{58.62} & \textbf{60.27} & \textbf{61.51} \\
\hline
\end{tabular}
\end{center}
\caption{
ResNet-18 Top-1 accuracy (\%) on the ImageNet validation set using 10\% labels and 30\%, 70\%, or 100\% unlabeled images.
}
\label{tab:part_unlabeled}
\vspace{-2mm}
\end{table}
\begin{table}
\begin{center}
\begin{tabular}{c|ccccccc|c}
\hline
\# labels & Supervised & Pseudolabels & VAT & VAT-EM & $S^4L$ & MT & \ShortName (ours) & LA* \\
\hline\hline
10\% & 44.7 & 48.2 & 45.8 & 46.2 & 46.6 & 46.4 & \textbf{50.4} & \multirow{2}{*}{48.3}  \\
\cline{1-8} %\hline
1\% & 36.2 & 41.8 & 35.9 & 36.4 & 38.0 & 31.6 & \textbf{44.6} \\
\hline
\end{tabular}
\end{center}
\caption{
ResNet-50 transfer learning Top-1 accuracy (\%) on the Places205 validation set using fixed weights trained on ImageNet using 10\% or 1\% of labels.
The numbers for models except ours, Mean Teacher's, and LA's are from~\cite{zhai2019s4l}. 
*: LA number is produced by us through training ResNet-50 with Local Aggregation algorithm~\cite{zhuang2019local} without any labels and then using the pretrained weights to do the same transfer learning task to Places205. 
Note that this reported number is lower than the corresponding number in~\cite{zhuang2019local}. 
This is because they reported 10-crop top-1 accuracy.
}
\label{tab:plc_part_labels_50}
\vspace{-2mm}
\end{table}
\begin{table}
\begin{center}
\begin{tabular}{c|ccccc}
\hline
Method & LS~\cite{zhou2004learning} & LP~\cite{zhu2002learning} & LP\_DMT~\cite{liu2018deep} & LP\_DLP~\cite{iscen2019label} & \ShortName (ours)  \\
\hline\hline
Perf. & 84.6 $\pm$ 3.4 & 87.7 $\pm$ 2.2 & 88.2 $\pm$ 2.3 & 89.2 $\pm$ 2.4 & 88.1 $\pm$ 2.3 \\
\hline
\end{tabular}
\end{center}
\caption{
Label propagation performance for different methods on subsets of ImageNet.
LS represents Label Spreading introduced by~\cite{zhou2004learning}.
LP represents Label Propagation introduced by~\cite{zhu2002learning}.
LP\_DMT represents the label propagation method used in DMT~\cite{liu2018deep}.
Similarly, LP\_DLP represents the method used in DLP~\cite{iscen2019label}.
For all methods, we randomly sample 50 categories from ImageNet and 50 images from each category.
For each category selected, we choose 5 images to be labelled.
For all methods, we use embedding outputs of our trained ResNet-50 with 10\% labels as data features.
The numbers after $\pm$ are standard deviations after 10 independent data subsamples.
}
\label{tab:label_prop}
\end{table}
\begin{table}
\begin{center}
\begin{tabular}{c|ccccccc|c}
\hline
Model & Top50woc & Top50 & Top20 & Top10 & Top5 & Top50wc & Top50lw & Top10wclw \\
\hline\hline
NN perf. & 52.43 & 54.37 & 54.82 & 55.42 & 55.44 & 55.46 & 56.27 & \textbf{57.54} \\
\hline
\end{tabular}
\end{center}
\caption{
Nearest-Neighbor (NN) validation performance (\%) for ResNet-18 trained using 10\% labels with various settings.
NN performances are lower than but correlate strongly with softmax classification performances reported in Table~\ref{tab:part_labels_18}.
Experiments labelled ``TopX'' involve training with $K=$X, optimizing only embedding loss, and updating the labels and confidence using the naive weighted $K$NN algorithm as defined in Eq.~\ref{equ:naive_weight}.
With these default settings, experiment with ``woc'' in the name have optimization loss not weighted by confidence.
``wc'' means the loss includes category loss.
``lw'' means using locally weighted probability for label propagation. 
Thus, ``Top10wclw'' indicates the \ShortName model. 
}
\label{tab:ablat}
\end{table}
%\subsection{ImageNet with different amount of unlabeled images}

\textbf{Transfer learning to Scene Recognition.}
To evaluate the quality of our learned representation in other downstream tasks besides ImageNet classification, we assess its transfer learning performance to the Places205~\cite{zhou2014learning} dataset.
This dataset has 2.45$M$ images total in 205 distinct scene categories.
We fix the nonlinear weights learned on ImageNet, add another linear readout layer on top of the penultimate layer, and train the readout using cross-entropy loss using SGD as above.
Learning rate is initialized at 0.01 and dropped by factor of 10 whenever validation performance on Places205 saturates.
Training requires approximately 500,000 steps, comprising two learning rate drops.
We only evaluate our ResNet-50 trained with 1\% or 10\% labels, as previous work~\cite{zhai2019s4l} reported performance with that architecture.
Table~\ref{tab:plc_part_labels_50} show that \ShortName again significantly outperforms previous state-of-the-art results.
It is notable, however, that when trained with only 1\% ImageNet labels, all current semi-supervised learning methods show somewhat worse transfer performance to Places205 than the Local Aggregation (LA) method, the current state-of-the-art unsupervised learning method~\cite{zhuang2019local}.

\section{Analysis} \label{sec:analysis}
To better understand the \ShortName procedure, we analyze both the final learned embedding space, and how it changes during training.

\begin{figure*}
\begin{center}
\includegraphics[width=\textwidth] {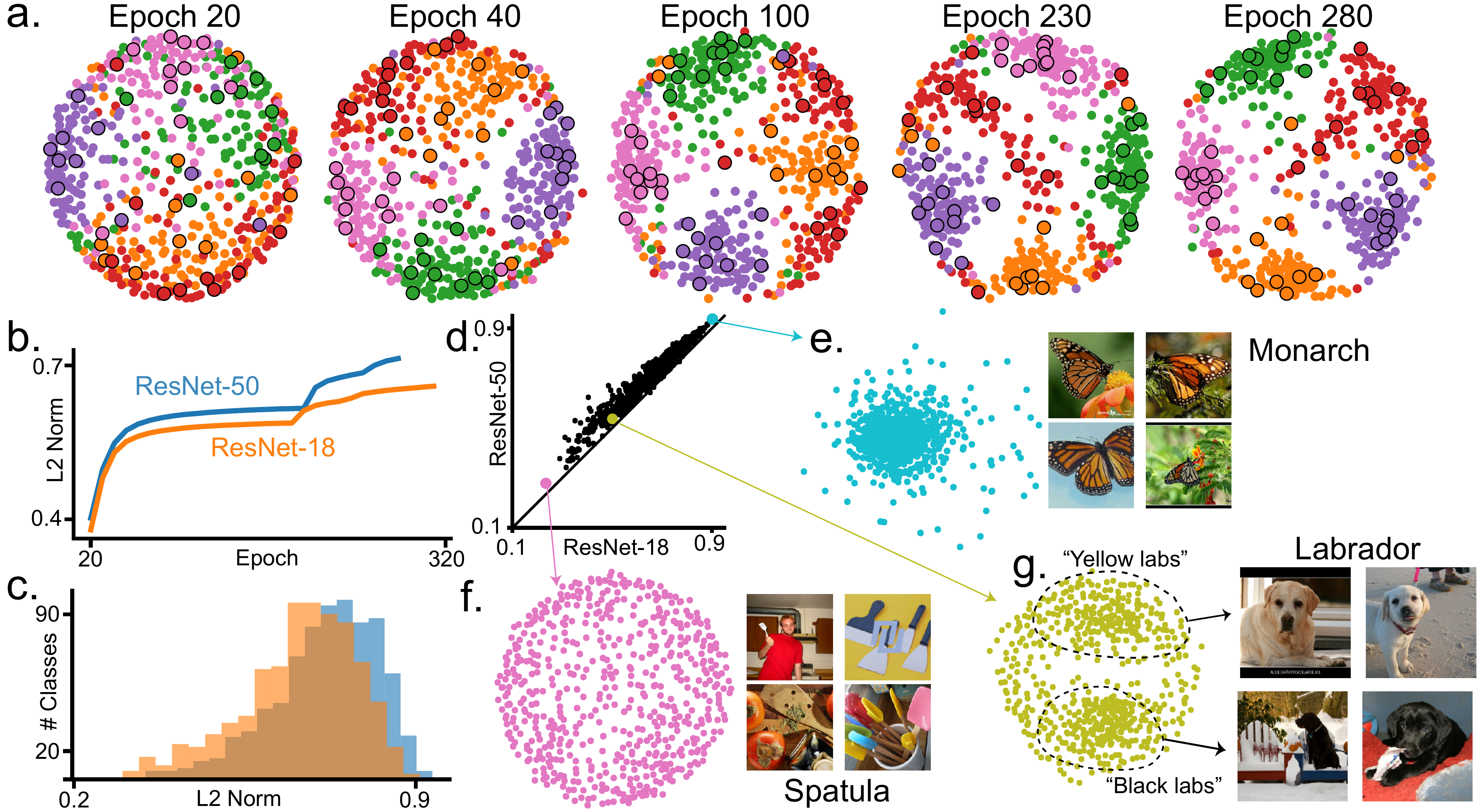}
\end{center}
\caption{
\textbf{a.} 2-dimensional MDS embeddings of 128-dimensional embedding outputs on 100 randomly sampled images from each of five randomly chosen ImageNet categories, from the beginning to the end of \ShortName training. 
Larger points with black borders are images with known labels.
\textbf{b.} Trajectory of cross-category average of L2-norms of category-mean embedding vectors for each ImageNet category. 
Sudden changes in trajectories are due to learning rate drops.
\textbf{c.} Histogram of the L2-norm metrics for each category, for fully-trained ResNet-18 and ResNet-50 networks. 
\textbf{d.} Scatter plot of L2-norm metric for ResNet-18 ($x$-axis) and ResNet-50 ($y$-axis). Each dot represents one category.
\textbf{e.-g.} MDS embeddings and exemplar images for images of ``Monarch'', ''Spatula'', and ''Labrador''. 
For each category, 700 images were randomly sampled to compute MDS embedding. 
In \textbf{g.}, exemplar images are chosen from the indicated subclusters.
}
\label{fig:embd_traj}
\end{figure*}

\textbf{Emerging clusters during training.} Intuitively, the aggregation term in eq. \ref{equ:loss_a} should cause embedding outputs with the same ground truth label, whether known or propagated, to cluster together during training. 
Indeed, visualization (Fig.~\ref{fig:embd_traj}a) shows clustering becoming more pronounced along the training trajectory both for labelled and unlabelled datapoints, while unlabelled datapoints surround labelled datapoints increasingly densely.
A simple metric measuring the aggregation of a group of embedding vectors is the L2 norm of the group mean, which, since all embeddings lie in the 128-D unit sphere, is inversely related to the group dispersion. 
Computing this metric separately for each ImageNet category and averaging across categories, we obtain a quantitative description of aggregation over the learning timecourse (Fig.~\ref{fig:embd_traj}b), further supporting the conclusion that \ShortName embeddings become increasingly clustered.

\textbf{Effect of architecture.}
We also investigate how network architecture influences learning trajectory and the final representation, comparing ResNet-50 and ResNet-18 trained with 10\% labels (Fig.~\ref{fig:embd_traj}b-d).
The more powerful ResNet-50 achieves a more clustered representation than ResNet-18, both at all time points during training and for almost every ImageNet category.

\textbf{Category structure analysis: successes, failures, and sub-category discovery.}
It is instructive to systematically analyze statistical patterns on a per-category basis.
To illustrate this, we visualize the embeddings for three representative categories with 2D multi-dimensional scaling (MDS). 
For an ``easy'' category with a high aggregation score (Fig.~\ref{fig:embd_traj}e), the \ShortName embedding identifies images with strong semantic similarity, supporting successful semi-supervised image retrieval.
For a ``hard'' category with low aggregation score (Fig.~\ref{fig:embd_traj}f), images statistics vary much more and the embedding fails to properly cluster examples together.
Most interestingly, for multi-modal categories with intermediate aggregation scores (Fig.~\ref{fig:embd_traj}g), the learned embedding can reconstruct semantically meaningful sub-clusters even when these are not present in the original labelling e.g. the ``labrador'' category decomposing into ``black'' and ``yellow'' subcategories.

\textbf{Comparison to global propagation in the small-dataset regime.}
To understand how \ShortName compares to methods that use global similarity information, but  therefore lack scalability to large datasets, we test several such methods on ImageNet subsets containing 50 categories and 50 images per category.
Table~\ref{tab:label_prop} shows that local propagtion methods can be effective even in this regime, as \ShortName's performance is comparable to that of the global propagation algorithm used in DMT~\cite{liu2018deep} and only slightly lower than that of DLP~\cite{iscen2019label}.

\textbf{Ablations.}
To justify key parameters and design choices, we conduct a series of ablation studies exploring the following alternatives, using: 
1. Different values for $K$ (experiments \textbf{Top50}, \textbf{Top20}, \textbf{Top10}, and \textbf{Top5} in Table~\ref{tab:ablat}); 
2. Confidence weights or not (\textbf{Top50woc} and \textbf{Top50}); 
3. Combined categorization loss or not (\textbf{Top50wc} and \textbf{Top50});
4. Density-weighted probability or not (\textbf{Top50lw} and \textbf{Top50}).
Results in Table~\ref{tab:ablat} shows the significant contributions of each design choice.
Since some models in these studies are not trained with softmax classifier layers, comparisons are made with a simple Nearest-Neighbor (NN) method, which for each test example finds the closest neighbor in the memory bank and uses that neighbor's label as its prediction.
As reported in \cite{wu2018unsupervised, zhuang2019local}, higher NN performance strongly predicts higher softmax categorization performance.

\section{Discussion}
In this work, we presented \ShortName, a method for semi-supervised deep neural network training that scales to large datasets.
\ShortName efficiently propagates labels from known to unknown examples in a common embedding space, ensuring high-quality propagation by exploiting the local structure of the embedding. 
The embedding itself is simultaneously co-trained to achieve high categorization performance while enforcing statistical consistency between real and pseudo-labels.
\ShortName achieves state-of-the-art semi-supervised learning results across all tested training regimes, including those with very small amounts of labelled data, and transfers effectively to other non-trained tasks.

In future work, we seek to improve \ShortName by better integrating it with state-of-the-art unsupervised learning methods (e.g. \cite{zhuang2019local}). 
This is especially relevant in the regime with very-low fractions of known labelled datapoints (e.g. $<$1\% of ImageNet labels), where the best pure unsupervised methods appear to outperform the state-of-the-art semi-supervised approaches. 
In addition, in its current formulation, \ShortName may be less effective on small datasets than alternatives that exploit global similarity structure (e.g. \cite{iscen2019label,liu2018deep}).
We thus hope to improve upon \ShortName by identifying methods of label propagation that can take advantage of global structure while remaining scalable. 

However, the real promise of semi-supervised learning is that it will enable tasks that are not already essentially solvable with supervision (cf.~\cite{izeki2019billi}).
Thus, an important direction for future work lies in applying \ShortName or related algorithms to tasks beyond object categorization where dense labeling is possible but very costly, such as medical imaging, multi-object scene understanding, 3D shape reconstruction, video-based action recognition and tracking, or understanding multi-modal audio-visual datastreams.

\clearpage
{\small
\bibliographystyle{ieee}
\bibliography{refs}
}

\end{document}